\documentclass[lettersize,journal]{IEEEtran}
\usepackage{amsmath,amsfonts}
\usepackage{algorithmic}
\usepackage{algorithm}
\usepackage{array}
\usepackage[caption=false,font=normalsize,labelfont=sf,textfont=sf]{subfig}
\usepackage{textcomp}
\usepackage{stfloats}
\usepackage{url}
\usepackage{verbatim}
\usepackage{graphicx}
\usepackage{orcidlink}
\usepackage{float}
\usepackage{cite}
\newcommand{\specialcell}[2][c]{%
  \begin{tabular}[#1]{@{}c@{}}#2\end{tabular}}

\begin{document}

\title{PReP: Efficient context-based shape retrieval for missing parts}

\author{
\IEEEauthorblockN{Vlassis Fotis \textsuperscript{1, 2} \orcidlink{0000-0002-1212-5500}},
Ioannis Romanelis \textsuperscript{1, 2} \orcidlink{0000-0002-2917-8705}, Georgios Mylonas \textsuperscript{2}, Athanasios Kalogeras \textsuperscript{2} ~\IEEEmembership{Senior Member,~IEEE}, Konstantinos Moustakas \textsuperscript{1} \orcidlink{0000-0001-7617-227X} ~\IEEEmembership{Senior Member,~IEEE}

\thanks{\textsuperscript{1}: ECE Department, University of Patras, Greece}
\thanks{\textsuperscript{2}: Industrial Systems Institute (ISI), Athena Research Center, Greece}

\thanks{Emails (in order): vfotis@ece.upatras.gr, iroman@ece.upatras.gr, kalogeras@isi.gr, mylonasg@athenarc.gr,  moustakas@ece.upatras.gr)}
\thanks{* The present paper has been developed as part of the RRREMAKER project, funded by the European Union 2020 Research and Innovation program under the Marie Sklodowska Curie - RISE grant agreement no 101008060, and the framework of H.F.R.I call “Basic research Financing (Horizontal support of all Sciences)” under the National Recovery and Resilience Plan “Greece 2.0” funded by the European Union -- NextGenerationEU (H.F.R.I. Project Number: 16469).}
}

\markboth{Journal of \LaTeX\ Class Files,~Vol.~14, No.~8, August~2021}%
{Shell \MakeLowercase{\textit{et al.}}: A Sample Article Using IEEEtran.cls for IEEE Journals}

\IEEEpubid{0000--0000/00\$00.00~\copyright~2021 IEEE}

\maketitle

\begin{abstract}
In this paper we study the problem of shape part retrieval in the point cloud domain. Shape retrieval methods in the literature rely on the presence of an existing query object, but what if the part we are looking for is not available? We present Part Retrieval Pipeline (PReP), a pipeline that creatively utilizes metric learning techniques along with a trained classification model to measure the suitability of potential replacement parts from a database, as part of an application scenario targeting circular economy. Through an innovative training procedure with increasing difficulty, it is able to learn to recognize suitable parts relying only on shape context. Thanks to its low parameter size and computational requirements, it can be used to sort through a warehouse of potentially tens of thousand of spare parts in just a few seconds. We also establish an alternative baseline approach to compare against, and extensively document the unique challenges associated with this task, as well as identify the design choices to solve them.

\end{abstract}


\section{Introduction}
\IEEEPARstart{S}{hape} retrieval is a fundamental problem in computer vision, even more so in recent years, where advances in scanning and ranging technology has lead to an abundance of 3D shape data to become available. In various domains such as industrial design, manufacturing \cite{b1}, robotics \cite{robotics1} as well as gaming, cultural heritage and archeology, being able to retrieve suitable shapes from a repository based on application-specific criteria is key. In this work we dive in to the much less studied problem of part retrieval and propose the first (to our knowledge) pipeline that can find a match for an object with missing parts, without being explicitly fed the information about which part is missing.

Without sacrificing generalizability and extensibility, for the rest of this work, we assume the following scenario; A maker hub is set up allowing users to bring their damaged objects and find replacement parts from a large, spare part warehouse. The goal is to facilitate circular economy chains, by giving a second chance to objects that are no longer of use due to the damage they have sustained. The objects are first scanned and converted to point cloud form, before being fed into a deep learning pipeline that finds suitable matches among potentially tens of thousands of spare parts in the warehouse, procured by disassembling other unsalvagable objects.

\begin{figure}[t]
\centerline{\includegraphics[width=3.35in]{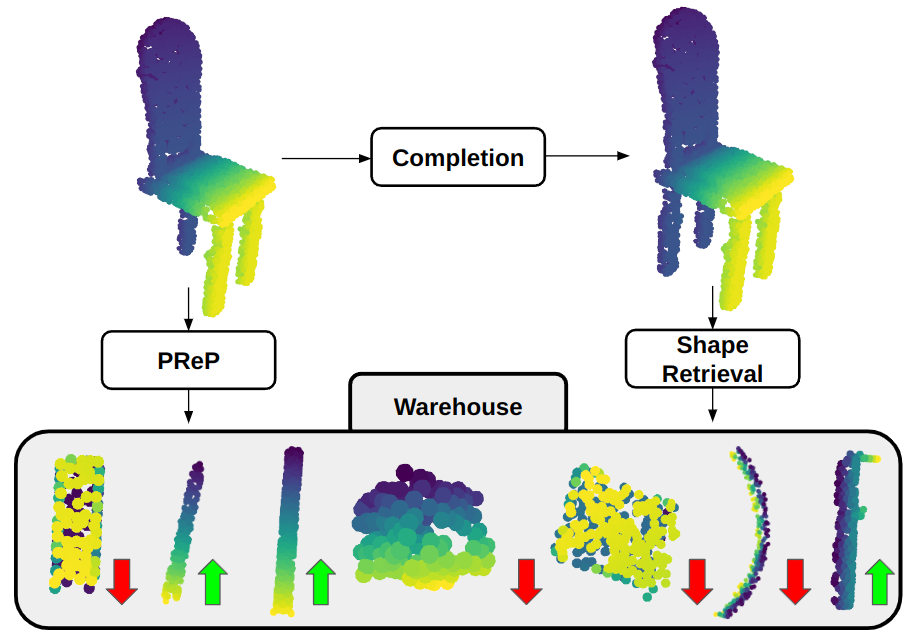}}
\caption{An overview of the proposed methods. An query object is missing a part and we need to find a replacement among a given database of parts. Using our proposed method, \textbf{PReP}, the matching part can be found only based on the context. By adding an additional completion step, standard shape retrieval techniques can be applied to retrieved a similar part.}
\label{fig:pipeline}
\end{figure}

Generally, deep learning-based works for shape retrieval work by encoding the query shape, as well as every other potential shape into latent space feature vectors. Then, a similarity measure between the feature vectors is applied, typically $\mathcal{L}_2$ distance or cosine similarity, and the matched objects are sorted in ascending order based on the distance value. The idea is that while it is difficult to match shapes due to uneven point density, noise and other factors, a deep neural network can generate a latent space in which similarity is encoded in the distance between feature points. The works then boil down to the architecture they use to tackle their specific problem, be it image, mesh, point cloud, cross modal etc, as well as the loss function they utilize to create the aforementioned feature space. The key difference between this problem and other related works in the shape retrieval domain is that the missing part is not actually available, and therefore it is not possible to scan it and find a similar one, having to rely solely on the context instead.

Our contributions are as follows:

\begin{itemize}
    \item We propose the first, to our knowledge, pipeline that performs shape retrieval on \textit{unknown parts} of the shape, whilst maintaining good computational efficiency and robust retrieval accuracy.
    \item We introduce several architectural novelties into this pipeline and extensively explain the challenges and design choices at each step, providing valuable insights.
\end{itemize}

\section{Related Work}
Shape retrieval is a long-standing problem in the field of computer vision that has received more attention in recent years due to the explosion of available data, easily accessible to both researchers and practitioners. It refers to the process of analyzing the characteristics of a specific object, represented in any form, be it 2D or 3D, and retrieving other objects with similar characteristics among a large database of objects. 

Shape retrieval challenges can be categorized according to the data they use as input and the data they try to retrieve. As such, 1D-to-2D \cite{LCFRCM, CMRDBR, MTCPAH}, 2D-to-2D \cite{ir1,ir2,ir3,ir4,ir5, SDHLSM, ADMLAI, RREVRV}, 2D-to-3D \cite{mm1,mm2,mm3,mm4,mm5,mm6,mm7,mm8, UCMGCN} and 3D-to-3D \cite{shrec2020} are all common areas of research. Recently, the focus of most works has shifted towards 2D-to-3D multimodal architectures in order to utilize the ever increasing data produced by various sensors. Matching 3D shapes from multiple views is a very popular approach, as it allows the models to acquire complete knowledge of objects, by assembling partial representations. Researchers utilize CNNs \cite{mm7,mm8}, N-Grams, \cite{mm6} as well as attention \cite{mm5} to extract and fuse features from multiple views. 

additonally, 3D shape retrieval from single image views is a much harder problem that has not gone unnoticed. In \cite{mm3}, the authors use modern contrastive learning techniques, instead of the standard metric learning methodology, to create a shared space. \cite{mm4} further investigates the problems of occlusions and unseen objects in single images. \cite{mm1} presents a novel framework for constructing a knowledge graph, which can be used to query with any kind of data.

Another problem that has been receiving attention lately is shape retrieval from abstract 2D sketches. In \cite{sketch3}, the authors propose various methods to retrieve 3D objects from 2D sketches of different artistic skill levels, by leveraging the QuickDraw \cite{sketch4} dataset. They approach the problem by creating a shared representation space by jointly training models that process sketches and 3D objects. \cite{sketch5} argues that a more effective approach would be to align the shapes and sketches in their common class label space instead, as a more effective way to bridge the gap between domains. In \cite{sketch6} the authors take a different approach; they train a model for 3D shape classification and use it as a teacher while a second model that learns 2D features from sketches becomes the student. They argue that this is a more effective way of learning a joint feature space, instead of training the two models simultaneously through a shared loss. \cite{sketch2} focuses on amateur-drawn sketches, in an attempt to democratize this process. They employ pivoting to address the lack of proper datasets for this task and fill the void between sketches and 3D shapes by including an intermediate image-based step. In \cite{sketch1}, the authors focus on the uncertainty aspect of this task, by attempting to remedy the effects of noisy or low-quality sketches in the data. They decouple the uncertainty from the rest of the retrieval task, and tackle it separately by injecting uncertainty information into the training of a classification model.

Niche tasks have also been attracting some attention. \cite{sdf} uses signed distance functions and emphasizes the need to perform retrieval for models that are not necessarily in the same scale or orientation as the query model. In \cite{shrec2020} they tackle the interesting problem of shape retrieval based on the surface pattern of the query model instead of the shape’s global structure. Promising results are presented using both geometric and learning-based methods.

Fewer works have tackled the problem of part retrieval. In \cite{b1}, the authors use a target object and assemble a lookalike by finding replacements for each of the object's components from a user defined database. They extract the parts through an optimization scheme before projecting them into a feature space and refining them using other decompositions as templates. In \cite{deform}, they perform 2D-to-3D retrieval but with an emphasis on parts. After finding a good initial match for the query image in the available 3D object repository, they deform the parts individually to achieve a greater similarity with the query object. 

All of these works assume that the parts in question are available. In our work we aim to specifically retrieve the piece that is missing, based only on the context of the shape.

\section{Methodology}

\begin{figure*}[h]
\centerline{\includegraphics[width=6.7in]{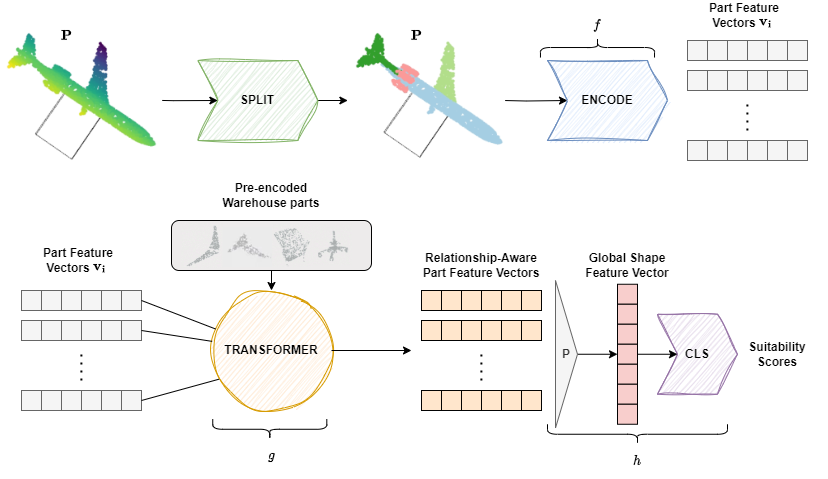}}
\caption{This figure presents an overview of our proposed pipeline. The input shape (in this case, a toy plane) is split into its individual parts manually or through a segmentation network and each part is encoded into a feature vector. The feature vectors are then fed into the transformer to model the relationships between them. At this stage the pre-encoded spare parts are also introduced to the transformer one at a time. For each spare part, the object is assessed as a whole, and a suitability score is provided by the trained classifier. (Each module is labeled with the corresponding mathematical symbol which is used to describe it in the text)}
\label{fig:pipeline}
\end{figure*}

All shape retrieval tasks require a way to quantify the suitability of every possible sample available in the retrieval database and then sort the samples according to the scores. A typical approach is as follows; a deep encoder is trained using metric learning principles, such as variants of triplet loss \cite{mll1, mll2, mll3, mll5, mll6, mll7}, in order to map similar samples into close points in the feature space. Once this is accomplished, the task of sorting samples based on suitability boils down to calculating their distance from the query shape in feature space, using some distance metric, often standard $L_2$ distance. 

This approach assumes that the query shape is available so that other samples can be compared against it. Nevertheless, in this case the part that is unavailable is precisely the one we are attempting to replace, so we need an alternate formulation. The proposed pipeline is summarized in figure \ref{fig:pipeline}. Let us assume that we are given point cloud objects $\mathbf{P}$ comprised of individual parts 

$$\mathbf{P} = \{\mathbf{p_1}, \mathbf{p_2}, ..., \mathbf{p_n} | \mathbf{p_i} \in \mathbb{R}^{N_i \times 3}\}$$

Let $\mathbf{v_i} = f(\mathbf{p_i}; \theta), \mathbb{R}^{N_i \times 3} \rightarrow \mathbb{R}^d$ be an encoder parameterized by a neural network with weights $\theta$, which maps the part point clouds to $d$-dimensional feature vectors. Additionally, let $s = h \circ g(\mathbf{v_1}, \mathbf{v_2}, .., \mathbf{v_M}; \psi), \mathbb{R}^{M \times d} \rightarrow \mathbb{R}$ be a secondary module, which takes as input $M$ part feature vectors and finally outputs a suitability score. Our objective is to replace a missing part $\mathbf{v_i}$ with a suitable spare part $\mathbf{\hat{v_i}}$. The encoder projects the parts into a high dimensional feature space, such that similar part geometries correspond to close feature points $\Vert \mathbf{v_i} - \mathbf{\hat{v_i}} \Vert < \epsilon$. Provided that $h \circ g$ is smooth, then $|s - \hat{s}| < \delta$, where $\epsilon, \delta$ are two arbitrarily small, positive numbers. 

We realize the individual components described above as follows:

\begin{itemize}
    \item $f$ is a deep pointnet-like encoder, which is pretrained to map similar geometries into close points in the feature space. The pointnet architecture guarantees that the parts can be processed in a permutation and cardinality invariant manner, which is an absolute necessity, as objects are composed of varying numbers of parts, and each part's geometry is represented by an arbitrary amount of points.
    
    \item Since the object's class is known even in the absence of $\mathbf{v_i}$, $h$ can be realized through a model trained for classification, wherein instead of outputting a single scalar the model outputs $K$ scores one for each class. During inference, we simply look at the output neuron that corresponds to the query object's class.

    \item Information-wise, $h \circ g$ is in charge of processing the part features and pooling them into an object-level feature vector, which is forwarded to the classification MLP. We implement $g$ using a shallow transformer to achieve this, since the attention modules will help encode the interactions between parts and provide a relationship-aware global embedding. During this stage, a pre-encoded potential replacement part $\hat{\mathbf{v_i}}$ is added in place of $\mathbf{v_i}$.
\end{itemize}

 An example test run of the pipeline will provide a deeper insight into its functionality:

A toy plane with a missing wing is given as input. The individual parts (such as body, tail and turbines) are encoded into the feature space and are processed further to output a classification score. If the missing wing was present, the 'plane' classification score would be high. The secondary module ($h \circ g$) is run continually, each time adding a different spare part to the mix. When a spare part with similar geometry as the original wing is encountered, the 'plane' classification score will be high again, signifying that a good replacement has been found. Consequently, this pipeline can be ran for every potential part candidate and the classification score can be used to compare each part suitability, given the current shape context.

In practice, the parts of an object are not necessarily given, but can be made available by using a trained segmentation network, or by having the maker separate them in a 3D processing software. The attention mechanism's memory and computational requirements scale quadratically with the number of parts, which in practice are very few (typically $<20$). Therefore, the transformer is able to evaluate hundreds of thousands of parts in just a few seconds. We make the reasonable assumption that the spare parts that are stored in a warehouse have been pre-encoded offline, so they do not impact the execution time during inference and as a result, the encoder can be made as big, parameter-wise, as needed.

\subsection{Training}

The pipeline is trained in two stages. First we want the encoder to assign close feature points to similar parts. Standard metric learning approaches recommend using the triplet loss, pioneered in \cite{mll1} to perform clustering of different viewpoints of face images. This technique has seen massive success, and further improvements in terms of sampling strategy \cite{mll2, mll3, mll5, mll6, mll7} as well as similarity measure \cite{mll4, mll8} have been proposed. Other variants \cite{mll9, mll10, mll11, mll12} have also been successful in similar tasks, reformulating the optimization objective significantly in order to obtain better convergence, learn more discriminative features or acquire tolerance to transformations. 

While the overwhelming success in the image domain might seem quite appealing, we argue that the harsh penalization of negative pairs might hinder the ability of the model to perform retrieval. This is because we would like to be able to retrieve parts that are geometrically similar but are not necessarily from the same class. Additionally, due to the continuous nature of point clouds - as opposed to the discrete nature of images - forming positive pairs through transformations increases the difficulty of the task dramatically. Instead, we propose a softer and computationally lighter training objective. 

First, the encoder is trained to learn to assign close feature points to parts that are from the same class e.g. chair legs. More formally, given a batch that contains parts $\mathbf{P} = \{p_i | i=0,\dots,N-1\}$ with corresponding part labels $l_i$ we minimize the following  function:

$$\mathcal{L} = \frac{1}{N^2} \sum_{i=0}^{N-1}\sum_{j=0}^{N-1} (S_{ij} - G_{ij})^2$$

$$\text{where }S = \mathbf{P} \cdot \mathbf{P}^T \text{ and } ||p_i|| = 1$$

\[ \text{and } G = \{g_{ij}\}, \text{ where } g_{ij} = 
\begin{cases}
1 & \text{if } l_i = l_j \\
0 & \text{otherwise}
\end{cases} \]

\begin{figure}[t]
\centerline{\includegraphics[width=3.4in]{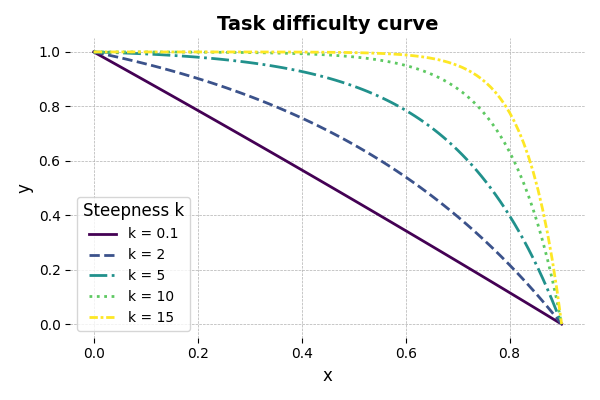}}
\caption{Visualization of $g$ for varying levels of steepness. The steeper the curve, the bigger the transition is from "highly similar" to "dissimilar" between objects. By gradually increasing the difficulty of the task the model becomes better at assigning feature points for parts based on their similarity.}
\label{fig:difficulty}
\end{figure}

In the above expression, $S$ is a dot product similarity matrix between each pair of parts, where all parts are first normalized to unit length. $G$ is a ground truth similarity matrix that is formed using the available part labels, forcing the parts of the same class to have high similarity and vice versa.

It is important to note that imposing this constraint on a model operating on whole objects may cause the model to overfit instead of learning any useful information, due to the significant differences in the objects' geometries. Nevertheless, our model operates on parts, which more often than not, share similar geometries when they belong to the same class. Consequently, the model learns to associate similar geometries through their labels. However, we found that the penalization of negative pairs resulted in a reduction in the number of retrieved matching parts from other classes, which is a highly desirable property. Due to this we need to relax the constraints further and incorporate shape similarity into the mix. We therefore replace the hard $0$s in the ground truth similarity matrix, with the following shape similarity metric:

$$
    g(d_{ij}) = \frac{e^{kd_h} - e^{kd_{ij}}}{e^{kd_{h}} - e^{kd_{l}}}
$$

Where $d_{ij}$ is the chamfer distance between two parts $\mathbf{P}_i$ and $\mathbf{P}_j$:

$$ d_{ij} = d(\mathbf{P}_i, \mathbf{P}_j) = \sum_{p \in \mathbf{P}_i}\min_{q \in \mathbf{P}_j} \|p-q\|^2_2 + \sum_{q \in \mathbf{P}_j}\min_{p \in \mathbf{P}_i} \|q-p\|^2_2$$

and $d_l$, $d_h$ are the chamfer distances of matched and mismatched parts respectively, averaged across a random subset of the dataset. All parts are normalized with respect to translation and scale before computing the chamfer distance. The function $g$ produces values close to 1 for shapes with high similarity (as measured by chamfer distance) and 0 otherwise. $k$ is a hyperparameter (see figure \ref{fig:difficulty}) which determines the steepness of the curve. We opted to use a scheduler for this hyperparameter, gradually increasing it during training to slowly increase the difficulty of the task, whilst facilitating the learning procedure. 

The second stage involves training the entire model for classification end-to-end. The idea is that the transformer must learn to predict which parts comprise an object of a specific class based \textit{solely} on the relationships between them. We want to keep the properties learned through the first stage of training intact \cite{expmae}, so we freeze the weights of the encoder during the second stage.

\subsection{Data preparation}

We evaluate our method on two popular benchmark datasets for segmentation, namely Shapenet-part \cite{shapenet} and PartNet \cite{partnet}. However, neither of those is designed for the task we are tackling, so they require a fair amount of data preparation. First, part groups such as chair legs are provided as a single entity and need to be separated. It is not possible to predict whether each sample will contain single parts or part groups, unless the splitting is performed manually. To this end, we leverage the DBSCAN clustering algorithm to separate the parts based on the density of points. 

This approach yields very good results on the high quality annotated PartNet but leaves a lot of problematic samples on ShapeNet, such as part groups that didn't end up splitting as well as parts with miniscule numbers of points that should not have been split but did. We discard parts that have a very low number of points, whilst being cautious about setting the threshold so as to avoid culling parts with small surface areas, which are bound to have fewer points than the rest. We view this as an opportunity to test the algorithm against imperfect data to evaluate the robustness of the proposed techniques.

After preprocessing and cleaning we create 2 datasets, one containing whole objects as well as part IDs for splitting, and one containing the processed parts. We will refer to these as the \textit{Items} and \textit{Warehouse} datasets respectively for the rest of this paper, for the shake of clarity. We feed the split parts to the model directly, omitting the segmentation network in favour of clarity and ease.

\section{Experiments}

\subsection{Retrieval Results}

\begin{figure*}[ht!]
\centerline{\includegraphics[scale=0.45]{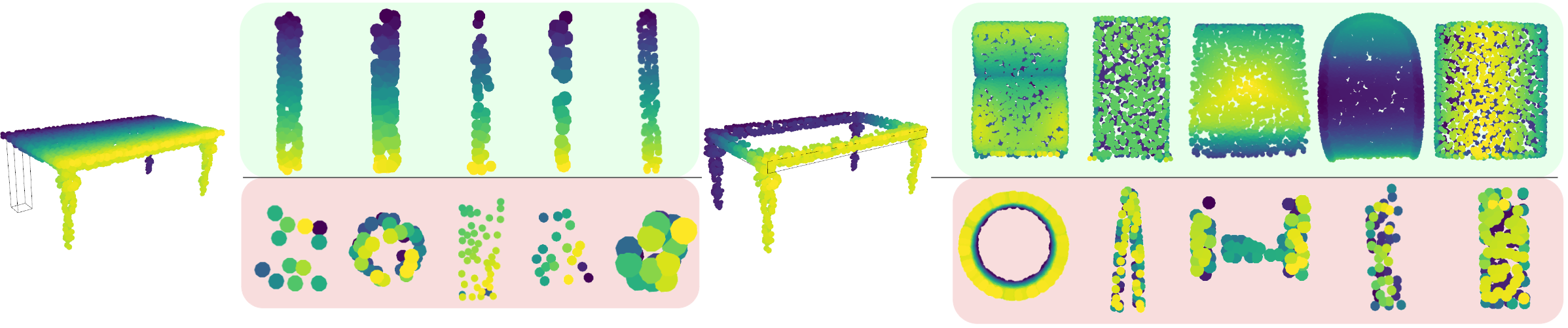}}
\caption{Retrieval results on the Partnet dataset. The same query table can be seen in two scenarios; one with a missing leg and one with a missing surface. The shapes on the top and bottom rows (color coded with green and red) represent the top and worst 5 matches respectively. Despite the various imperfections deriving from improper clustering the model is able to pick out fitting replacements. The colormap represents the normalized z-coordinate of each point and assists in understanding the object's geometry. \vspace{1cm}}
\label{fig:results_partnet}
\end{figure*}

\begin{figure*}[ht!]
\centerline{\includegraphics[scale=0.45]{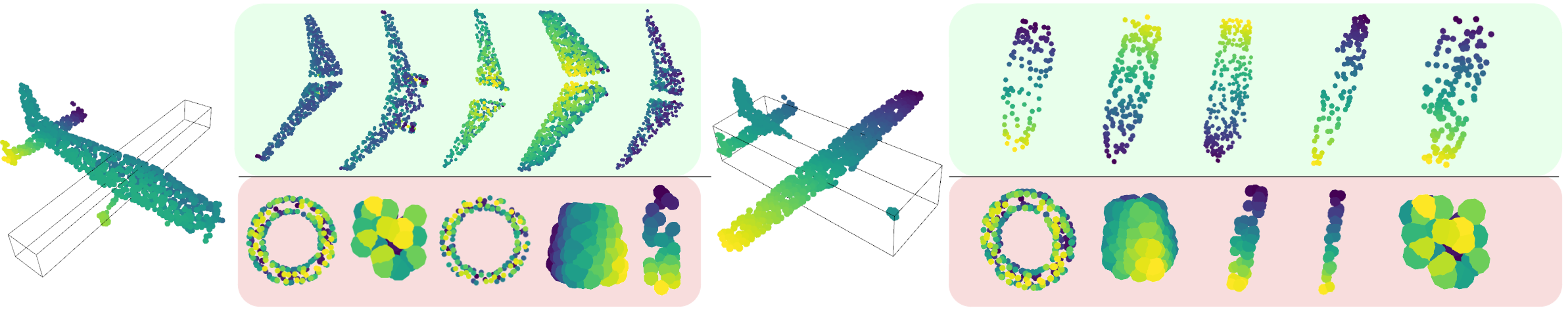}}
\caption{Retrieval results on the ShapeNet-part dataset. The same query plane can be seen in two scenarios; with missing wings and with missing fuselage. The parts on the top and bottom rows represent the best and worst matches respectively. The model is able to achieve good retrieval quality, despite the data being undersampled and the parts often being mislabeled.}
\label{fig:results_shapenet}
\end{figure*}

To evaluate our model we perform the training exactly as described in the previous section, and select several classes from Shapenet according to their popularity among research papers for demonstration purposes. For PartNet, we select classes with a reasonable amount of parts, making sure that these particular classes contain enough training samples so that the model can be trained properly. After training, we perform the retrieval process as follows: we randomly select a query shape from the Items dataset, split it and encode its parts, before running the transformer module repeatedly for all parts in the warehouse dataset. For each part used, we check the Nth output of the classification head, where N corresponds to the class of the query shape. We save these, as they are the scores that will determine the suitability of the spare parts. Finally, the parts are sorted according to their score, and the top-K are to be presented to the user of the application.

As can be seen in Figure \ref{fig:results_shapenet}, despite the rather extreme variation in the number of points, as well as the presence of unwanted parts due to mistaken clustering, the pipeline is able to select meaningful replacements. The same can be said in the case of PartNet \ref{fig:results_partnet} with its significantly more dense shapes and better annotation. We added four more layers to our pointnet and quadrupled the feature dimension to allow the encoder to process the PartNet shapes effectively. No changes were made to the relationship module, allowing the model to process the entire warehouse of spare parts in $\approx 13$ seconds on both datasets. It should be noted that the actual scale of the parts is not taken into consideration, on account of the normalization step that happens before feeding the parts to the model. The matching and retrieval depends solely on the geometry. A characteristic example of this can be seen in \ref{fig:results_shapenet}b, where the plane's fuselage is matched with not only other parts of its kind, but also with plane turbines, due to their similar geometry.

\subsection{Baseline}

Because the problem we tackle in this paper is novel, there are currently no other works -to the best of our knowledge- that could be used for fair comparative evaluation. As such, we establish a baseline method for this task, based on the simplest idea that might occur when studying this particular problem. Since the part that we are trying to replace is unavailable, we can use a trained completion model to fill the gap. Standard retrieval approaches can then be applied using the generated part as a query. 

To this end, we test 3 well-known completion networks, which we specifically select to have varying degrees of performance. This way, we can assess how the completion quality affects the retrieval end result. For evaluation we use the \textit{items} dataset for both approaches, discarding a single random part each time. For the completion approach, since it is common for models to output points even in areas where nothing is missing, we assume that the end user can specify an area of interest, where only points inside it are taken into account. To accomplish this automatically, we simply consider the bounding box of the missing part (which naturally will not be available in practice), but scale it up by a small factor to account for user error. We then isolate the points inside it and perform retrieval in 2 ways; by comparing the euclidean distance in feature space and by comparing the chamfer distance directly. We measure the chamfer distance between the retrieved and the original part and summarize the results in table \ref{t:cd}. While chamfer distance does indeed provide a baseline estimate of how the methods perform, the numbers themselves should be taken with a grain of salt, as it is not a suitable metric for this task, albeit the best one available. 

\begin{figure*}[htbp]
\centerline{\includegraphics[width=6.7in]{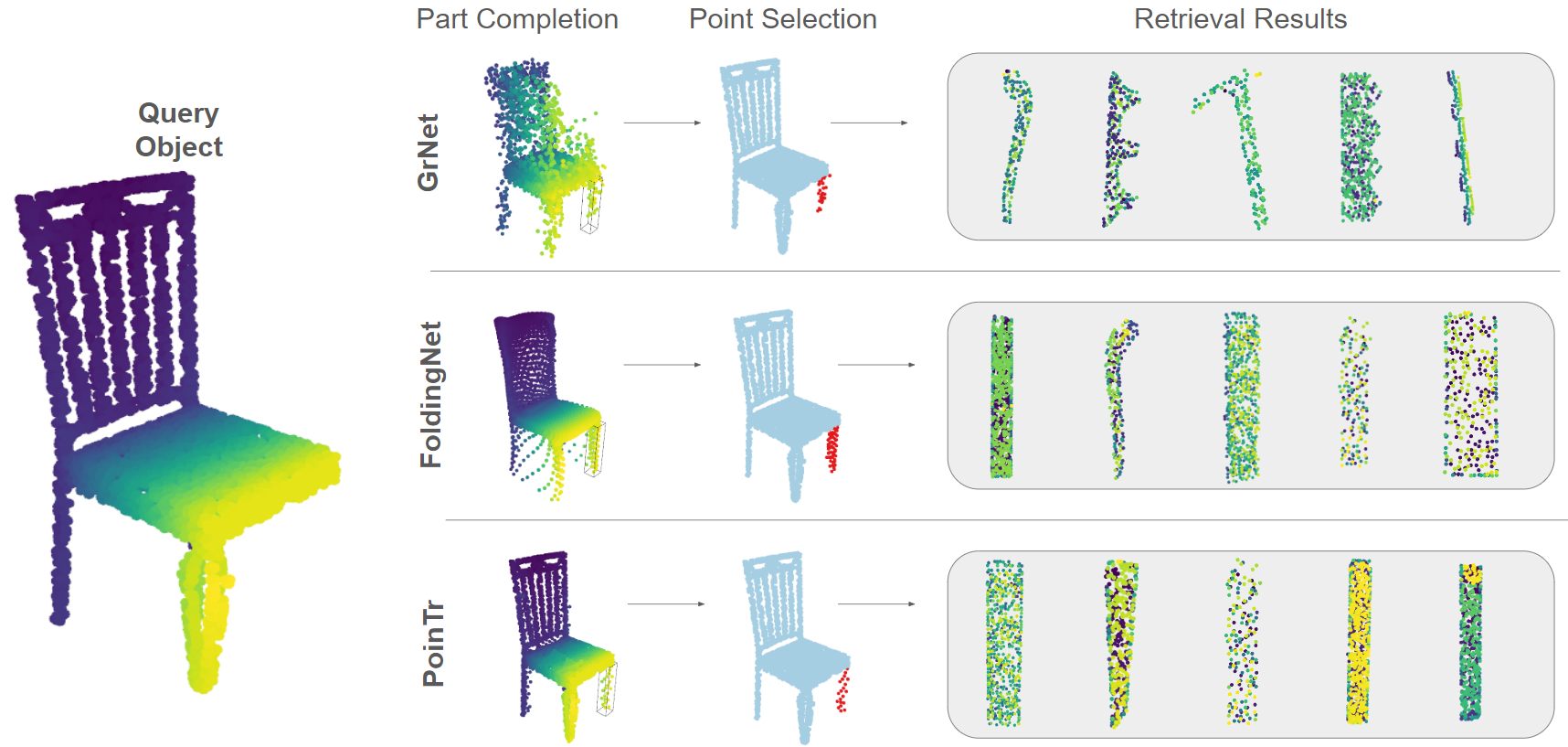}}
\caption{Visualization of the results produced by the baseline approach. The query model on the left is completed using 3 different models (from top to bottom): a) GrNet \cite{comp4},  b) FoldingNet as presented in \cite{comp2} and utilized for completion in 
\cite{comp3}, c) PoinTr \cite{comp1}. In cyan and red we visualize the generated points that are within the bounding box and are used for retrieval. On the right the top-5 matches are presented for each retrieval attempt. The retrieval was performed in point space, using chamfer distance.
}\label{fig:results_completion}
\end{figure*}

Completion and retrieval results are also visualized in Figure \ref{fig:results_completion}. The results are generally satisfactory, however as the completion quality worsens, the matched parts have noticeable issues, such as shape irregularities, protrusions as well as negligible thickness. Particularly, the almost non-existent thickness is a trait that helps achieve very low chamfer distance, as verified by the total score of each method.

\begin{table}
\caption{Comparative evaluation}
\label{t:cd}
\begin{tabular}{l|c c c}
Pipeline & \specialcell{Chamfer Distance \\ $(\cdot 10^{2})$} & \specialcell{Time / sample \\($\cdot 10^4$ s)} \\
\hline
 PoinTr (CD) & \underline{0.09} & 3.6\\
 GrNet (CD) & 0.14 &  3.6 \\
 FoldingNet (CD) & 0.10 & 3.6 \\
 \hline
 PRep & \textbf{2.13} & \textbf{2.8} \\
 FoldingNet (Enc) & 4.11 &  \underline{2.3} \\
 GrNet (Enc) & 4.36 & 2.3 \\ 
 PoinTr (Enc) & 3.81 & 2.3 \\ [1.2ex]
\hline 
\multicolumn{4}{p{210pt}}{
\textit{CD} and \textit{ENC} denote the way in which the samples are retrieved from the database; by comparing the chamfer distance or encoding the part and using comparing the euclidean distance respectively. Naturally, in the cases where the parts are retrieved using chamfer distance the average distance is lower than \textit{PRep}. (The experiments are ran on a single RTX 3090 gpu)} \\ 
\end{tabular}
\end{table}

This approach, despite being simplistic, does not come without its merits. Specifically, the "split" stage in PReP becomes obsolete, potentially saving compute power or the user's time that would otherwise be spent in highlighting the different parts. A strong downside, however, is that the quality of retrieval results depends \textbf{heavily} on the completion quality and any mishaps during this process, such as misplaced points or inaccurate geometry representations, will propagate through the pipeline and affect the retrieval quality. Despite evident advances in point cloud completion, the task remains relatively challenging. Oversampled and undersampled objects, as well as point clouds that have been affected by specific patterns of noise or artifacts have a sizable impact on the geometry being represented. 

In addition, completion networks rely heavily on class information to perform the task, some even requiring to be trained on a single class at a time. Consequently, encoding the underlying structure of objects of multiple classes requires an enormous amount of data, and an adequately sized model. On the contrary, while the classification task requires scaling up to deal with an increased number of classes, the task is significantly easier than completion and does not demand exorbitant amounts of data. 

\subsection{Multiple missing parts}

\begin{figure*}[ht!]
\centerline{\includegraphics[width=6.7in]{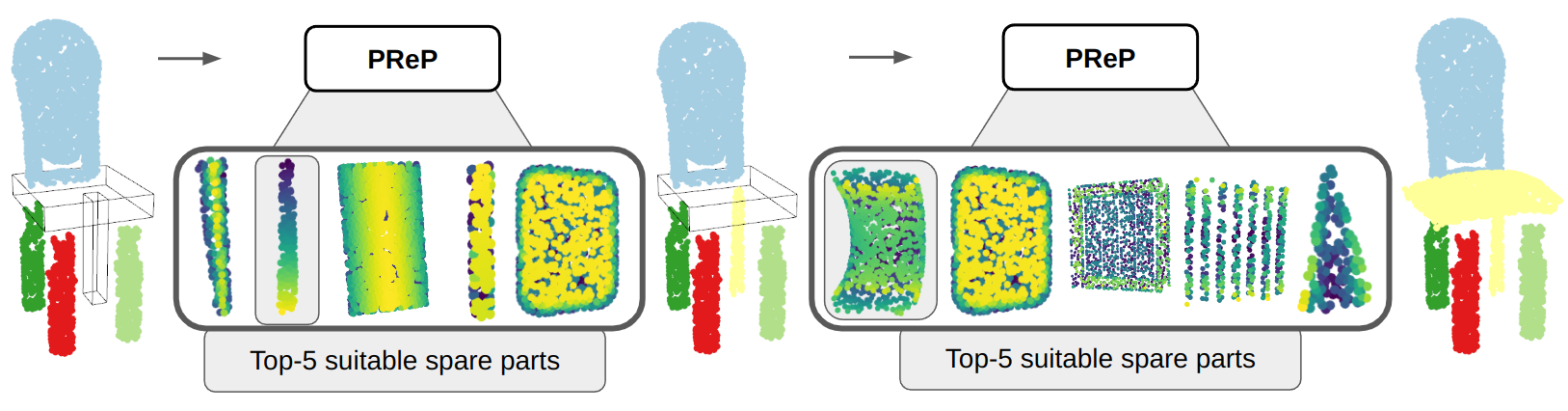}}
\caption{Multipart retrieval example. The query chair is missing two parts; a seat and a leg. We visualize the process of matching replacement parts, selecting an appropriate replacement and repeating this process until the object is fixed. PReP provides a list of viable replacement part candidates, from which the user will select the one that better suits his situation and/or tastes. More results can be found in the appendix.}
\label{fig:results_multiretrieval}
\end{figure*}

We extend the pipeline to tackle objects that have multiple missing parts by introducing recurrency in the retrieval process. After feeding the query object to the model and scoring the spare parts we add the selected replacement to the query object and repeat this process until it is complete. The parts are centered and normalized with respect to scale during the retrieval process, to make it easier for the model to find an appropriate match. Each time a replacement is found, it is translated to where the missing part is supposed to be and aligned using the principal axes, before beginning the next iteration to find the next replacement part. During this stage, we wish to observe whether the highest scoring parts belong to a singular class or from multiple, depending on which parts are missing. As can be seen in Figure \ref{fig:results_multiretrieval}, when two different parts are missing, replacements from both classes appear in the top matches. Respectively, when the replacement is found, the top matches mostly include parts from the remaining missing class, as intended. Note that as the replacement leg is added to the object, the suggestions for the seat do not necessarily remain the same, but the geometry requirements are satisfied.\\

\begin{figure}[t!]
\centerline{\includegraphics[width=3.35in]{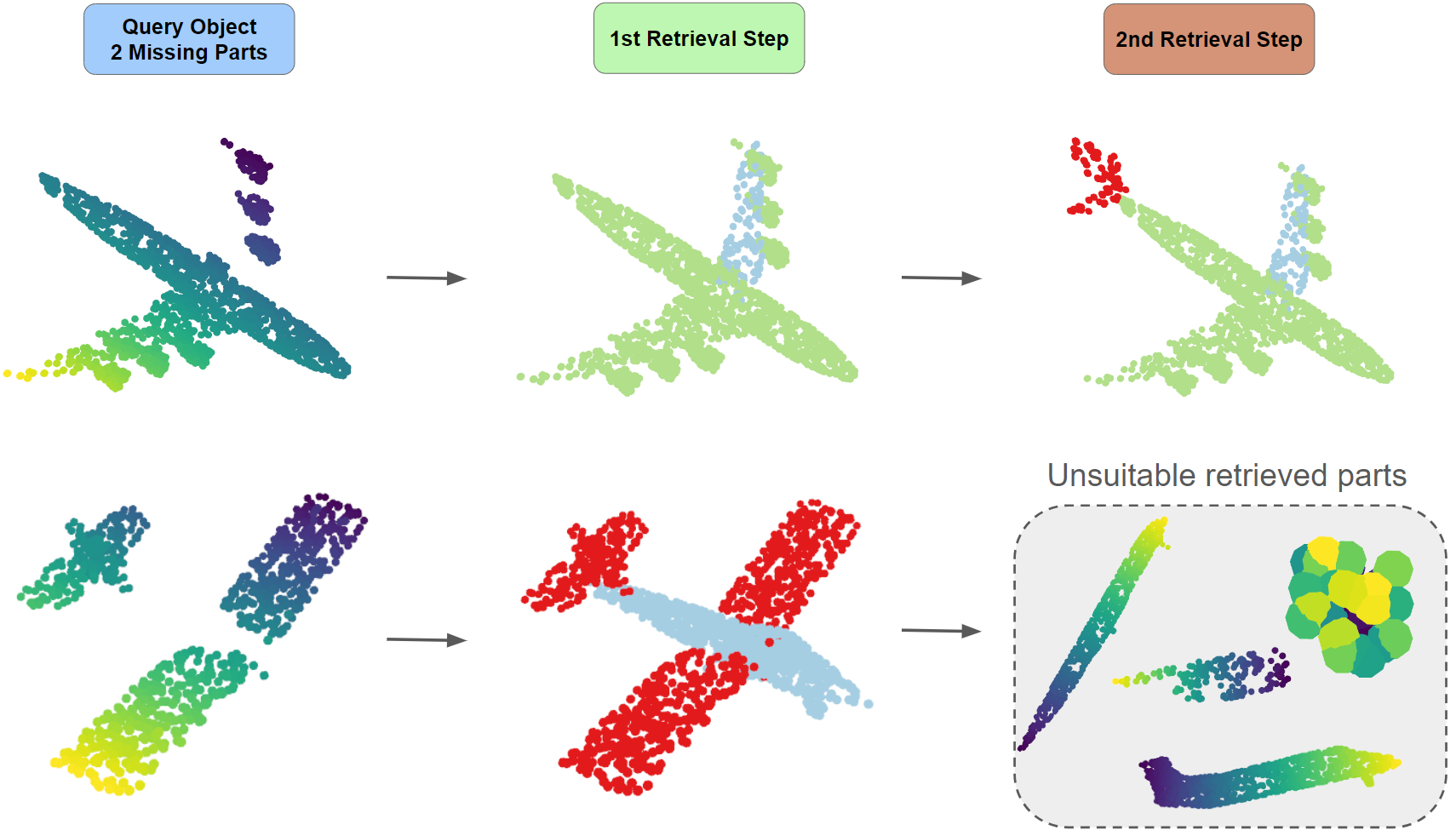}}
\caption{Additional results on multi-part retrieval on the shapenet dataset. On the left column we can see the objects provided as input that are missing two parts each. During every retrieval step, the added part shown in the figure has been selected among the top matches provided by PReP. The first and second retrieved part is displayed in cyan and red respectively.}
\label{fig:problematic_case}
\end{figure}

 Let us also provide some interpretability and deeper insights into the model's functionality as well as its limitations. First and foremost, this approach is limited by the capacity of the classifier. Once the shape has been stripped to the point where its class is not predicted correctly, the suitability scores will be unpredictable and the matched parts will not necessarily fit the context. In fact, this goes both ways; in Figure \ref{fig:problematic_case} the toy plane on the second row originally had two sets of wings. Upon retrieving a plane body during the first retrieval step, the classifier recognizes the shape as a full plane and produces a confident guess. The spare parts provided in the second retrieval step simply add noise to the already high classifier output, resulting in unsuitable matches. This is in contrast to the toy plane in the first row, where the object is indeed recognized as a plane, but the classifier's guess becomes more and more confident during each retrieval step as the replacements for the missing parts are found.

A tertiary, rather minor problem is that matches will tend to worsen as more parts are removed. This is because the part combinations that the model receives as input become increasingly "odd" compared to its training data. Nevertheless, this problem is solely attributed to the capacity of available point cloud processing models and their ability to understand geometric shapes regardless of sampling. 

\section{Discussion}

We have proposed \textit{PReP}, a pipeline that can perform part retrieval based on shape context, a significantly more difficult problem than traditional shape retrieval. \textit{PReP} is lightweight enough to sort through a repository of tens of thousands of spare parts in merely a few seconds. It owes its quality and speed to the creative usage of classification and the simplicity thereof. Additionally, we established a baseline approach based on completion and then retrieval of the missing part. While the latter will often yield better results in the "chamfer distance sense", it should be noted that chamfer distance is not really a suitable metric for this task, albeit the best one available. In actuality, suitable replacements do not necessarily adhere to this lowest NN distance rule, but require a broader level of geometry understanding which can only be provided by \textit{PReP}. The strict nature of chamfer distance not only limits the pool of potential replacements, but also increases sensitivity to errors (as in the example of Figure \ref{fig:results_completion}). Things are complicated even further when geometries become more complex. Nevertheless, the baseline still remains a feasible way to tackle this problem, so long as the diversity in the data allows for a proper completion network to be trained and deployed. 

\section{Future work}

Arguably, the most difficult aspect of this task is training  the encoder so that it can bring similar samples close in feature space. Due to the complicated nature of the data, with arbitrary number of parts in each object and arbitrary number of points per part, the encoding process involves a lot of masking and padding operations. This is not necessarily compatible with all point cloud processing models available and the selection of the best model requires a lot of attention and care. For the sake of simplicity we used a PointNet, which is hardly one of the best encoders available in the literature. In future work we plan to investigate the appropriate choice of the encoder further, potentially even taking the scale of the part into consideration.

\appendix[]

\section*{Data}

Care has been taken to keep the data used as close to other publications that utilize the same datasets, but making it relevant and fitting for this specific task. On ShapeNet-Part we simply iterate through the dataset and save its part individually, retaining information about the object class it originated from and the part class it belongs to. The same is done for PartNet, but each part cluster is split into subparts. The number of clusters is determined automatically by the DBSCAN algorithm, which works perfectly for the majority of the samples, but creates noisy, non-existent parts in degenerate cases, which unfortunately pollutes the dataset with unwanted samples. 

While the objects in both datasets are canonically oriented (e.g. all chairs face the same direction), the individual parts they comprise are not. Naturally, this would affect the Chamfer distance calculation so we remedy this problem by aligning the parts using principal component analysis. For elongated parts (such as chair legs, cylinders etc) this amounts to the direction of the sample itself which is the principal component with the largest eigenvalue. For planar surface parts we take the principal component with the smallest eigenvalue instead, which corresponds to the orientation of the surface. Each of these vectors is saved as metadata along with the part in question. During runtime, when comparing two parts together, we align these vectors and apply the same rotation matrix to one of the point clouds.

\section*{Model and Training}

As mentioned in the main paper the model is trained in two stages; first the encoder with the proposed metric learning objective, and afterwards the entire model for classification. We perform the first stage of training for 100 epochs, requiring approximately 8 hours on an RTX 3090 gpu. The classification stage is much lighter, trained for only 20 epochs, over a span of approximately 40 minutes. Regarding our training parameters, we used the Adam optimizer for both tasks along with cosine annealing with warmup for learning rate scheduling.  

Further details are required in order to understand the relationship module's functionality. The transformer takes as input encoded parts and outputs feature vectors of the same dimensionality by injecting information about the relationships between parts. Afterwards, a pooling operation is applied to convert the part feature vectors into a global shape feature vector, which is then provided as input in the classification MLP. Two problems arose in the early stages of this design:

First, the pipeline would output a very high score for a specific class, despite the input missing a part or two. This was primarily caused by the presence of max pooling. Distinctive features of a particular class would make their way into the classification MLP due to the max pooling operation, ignoring the fact that vital parts were missing. We remedied this problem by employing a classification token instead of max pooling.

The second issue, closely related to the first, is that the presence of extra parts or the absence thereof would not be appropriately reflected in the score. This is precisely the reason we decided to perform the training in two stages, first equalizing similar parts and then training for classification without adjusting the encoder. This separation of the encoder from the rest of the model allows the classification to be based only on the parts present, and not on changing the feature representation to make classification easier. In the ideal case, where for the same type of part the same exact feature vector is assigned, there needs to be a way for the transformer to differentiate between them. In \cite{assembly1} they encounter this problem in shape assembly, which they tackle by assigning a unique instance embedding to each part. We opt to use a similar approach, but we simplify and generate a unique embedding from the centroid of the part. By concatenating this embedding to each part feature vector we manage to kill to birds with one stone, as not only does it allow the transformer to distinguish between parts with similar geometries, but also helps the transformer understand the location of said parts.

\section*{Additional Results}

\begin{figure}[ht!]
\centerline{\includegraphics[width=3.35in]{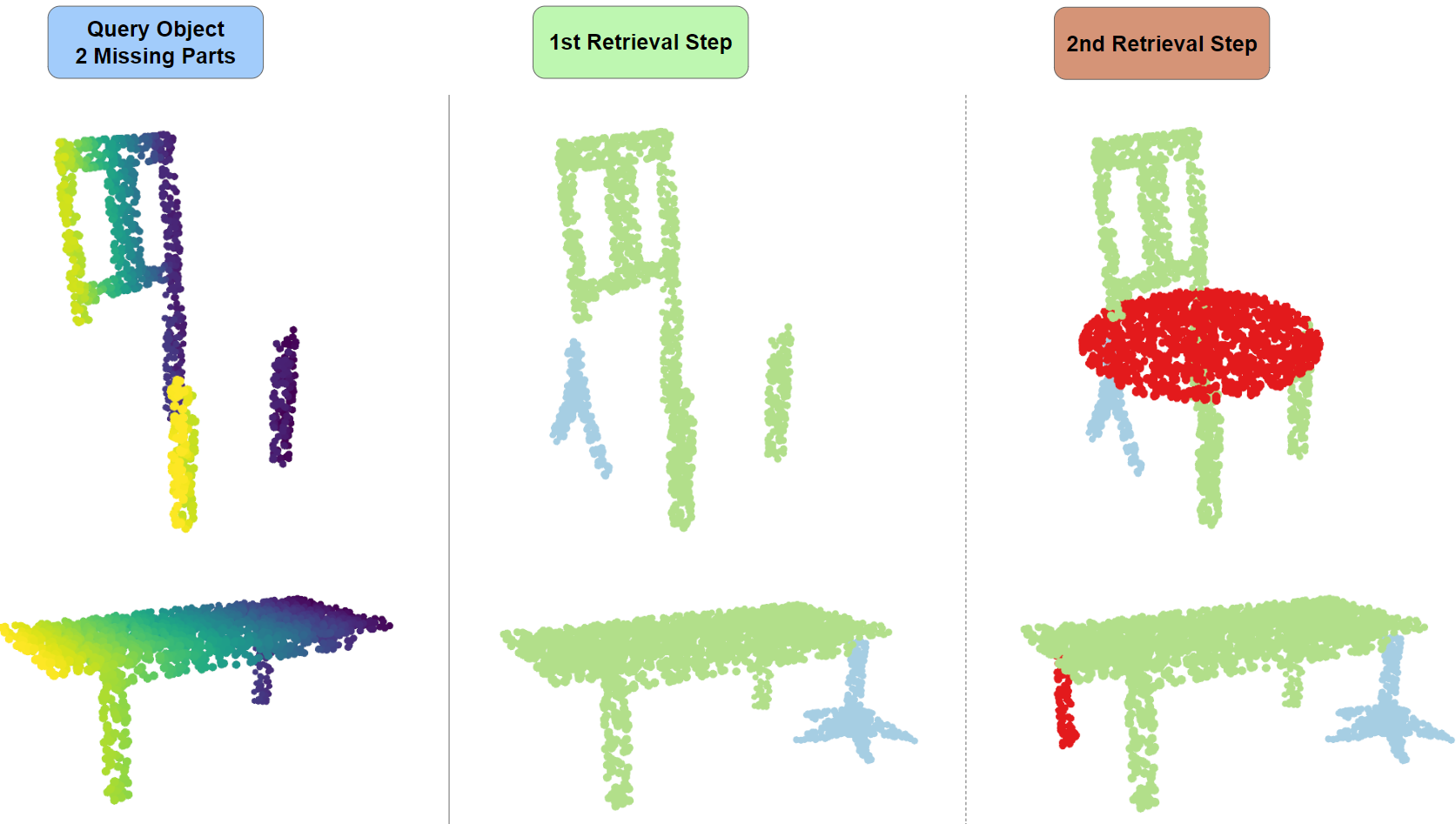}}
\caption{Additional results on multi-part retrieval on the shapenet dataset. On the left column we can see the objects provided as input that are missing two parts each. During every retrieval step, the added part shown in the figure has been selected among the top matches provided by PReP. The first and second retrieved part is displayed in cyan and red respectively.}
\label{fig:shapenet_appendix}
\end{figure}

\begin{figure}[ht!]
\centerline{\includegraphics[width=3.35in]{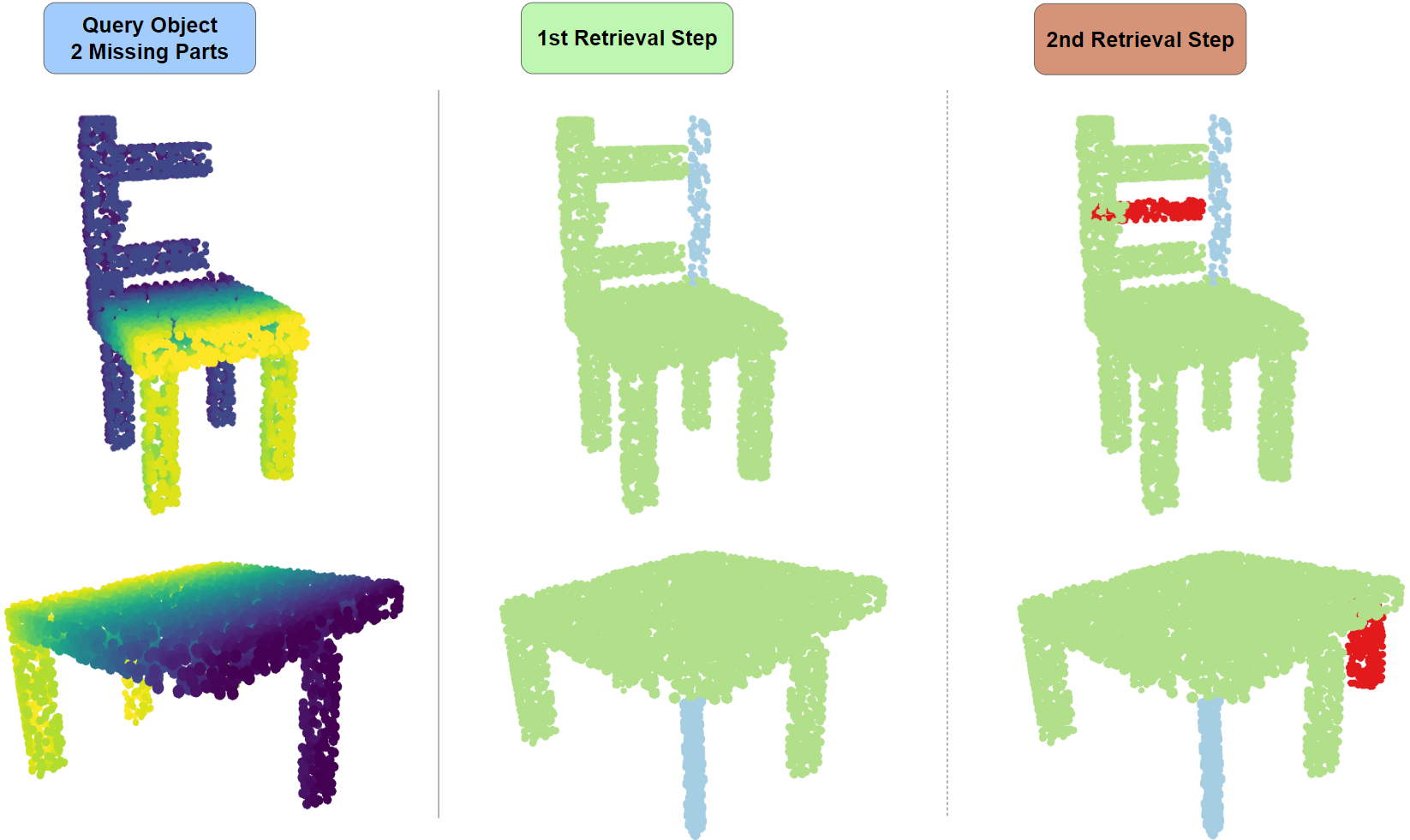}}
\caption{Additional results on multi-part retrieval on the partnet dataset.}
\label{fig:partnet_appendix}
\end{figure}

In figures \ref{fig:shapenet_appendix} and \ref{fig:partnet_appendix} we present additional results for multi-part retrieval. We would like to draw the reader's attention to the chair and table examples of figure \ref{fig:shapenet_appendix}, wherein the retrieved parts do not necessarily fit the query object's aesthetic, but do provide the required functionality, same as the object's original parts. This is a useful property in the scenario that we have assumed, because there is the option of physically processing the parts further, according to the user's needs, providing even greater variety of potential replacements.


\vspace{-2cm}

\begin{IEEEbiography}[{\includegraphics[width=1in,height=1.25in,clip,keepaspectratio]{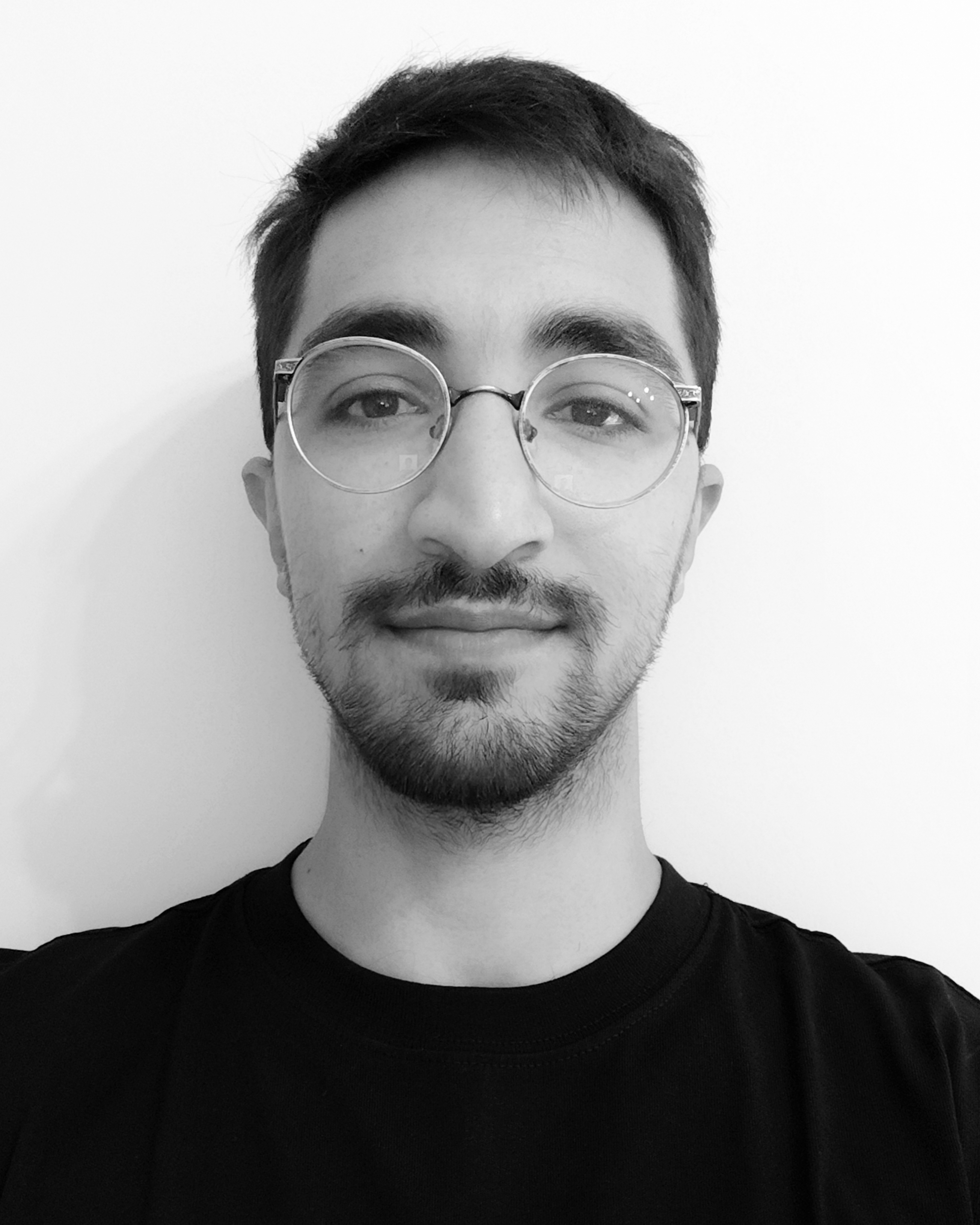}}]{Vlassis Fotis}
received his Electrical Engineering and Computer Technology degree in 2021 at the university of Patras. During the same year he joined the VVR lab as a PhD candidate. His main research interests include (but are not limited to) computer vision, theoretical deep learning, 3D scene understanding and geometry processing.
\end{IEEEbiography}

\vspace{-2cm}

\begin{IEEEbiography}[{\includegraphics[width=1in,height=1.25in,clip,keepaspectratio]{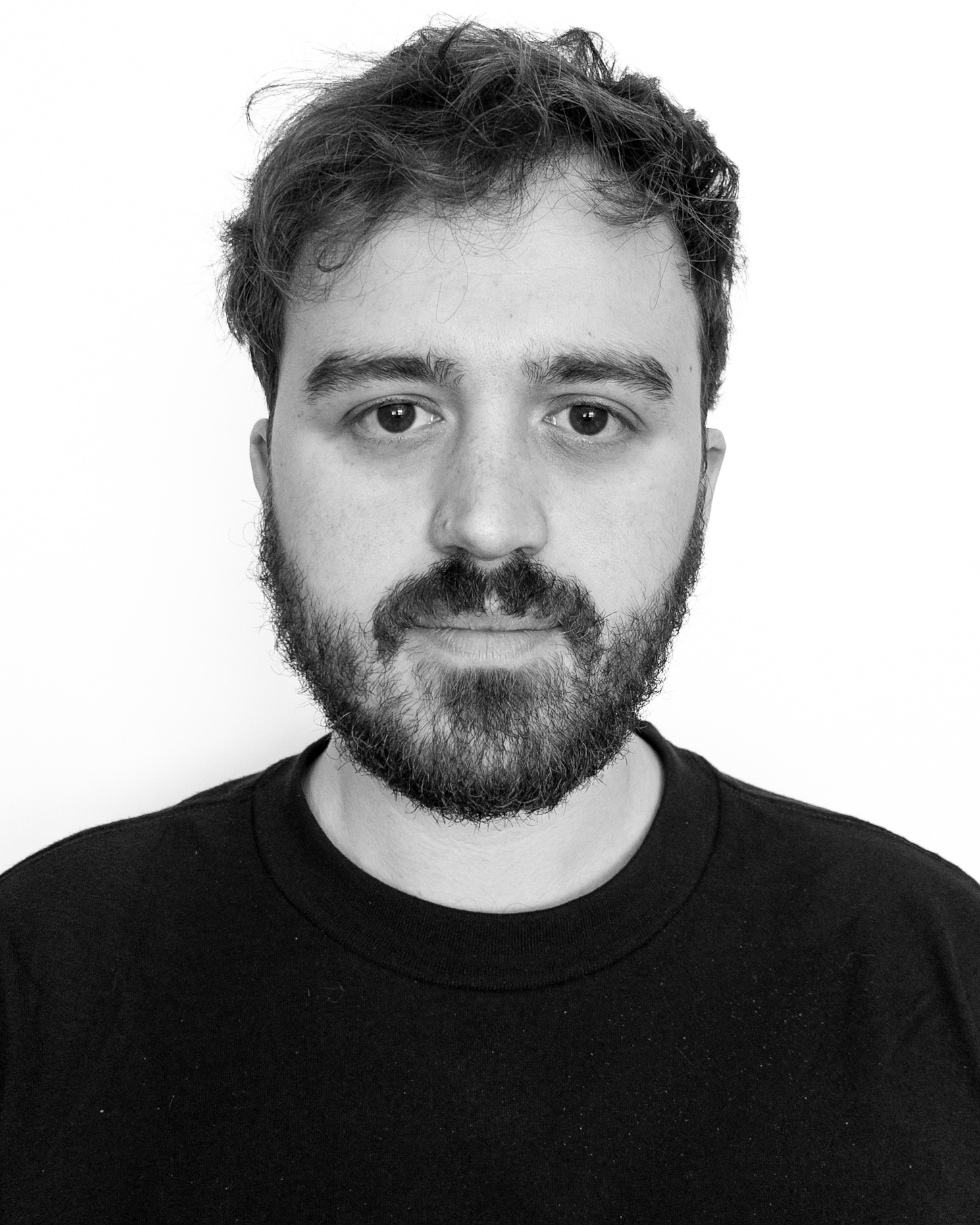}}]{Ioannis Romanelis}
received his Electrical and Computer Engineering diploma in 2021 at the University of Patras. During the same year, he enrolled for a PhD at the same department under the supervision of Professor Konstantinos Moustakas and
joined the Visualization and Virtual Reality (VVR) group. His main research interests include computer vision, deep learning, point cloud processing, 3D scene understanding, and explainable
AI.
\end{IEEEbiography}

\vspace{-2cm}

\begin{IEEEbiography}[{\includegraphics[width=1in,height=1.25in,clip,keepaspectratio]{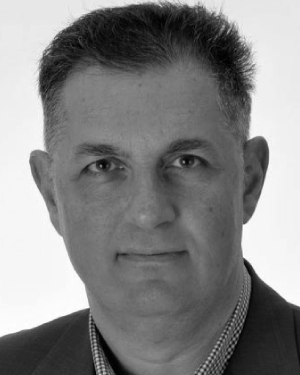}}]{Athanasios Kalogeras}  (Senior Member, IEEE) received the Diploma degree in electrical
engineering and the Ph.D. degree in electrical and computer engineering from the University of
Patras, Greece. He has been with the Industrial Systems Institute, ATHENA Research and Innovation Center, since 2000, where he currently holds a position of the Research Director. He has worked as an Adjunct Faculty at the Technological Educational Foundation of Patras. He has been a Collaborating Researcher at the University of Patras, the Computer Technology Institute and Press “Diophantus,” and the private sector. His research interests include cyber- physical systems, the Industrial IoT, industrial integration and interoperability, and collaborative
manufacturing. Application areas include the manufacturing environment, critical infrastructure protection, smart buildings, smart cities, smart energy, circular economy, health, and tourism and culture. He has served as a program committee member for more than 30 conferences and as a reviewer in more than 40 international journals and conferences. He has been a Postgraduate Scholar of the Bodossaki Foundation. He is a member of the Technical Chamber of Greece. He is a Local Editor in Greece of ERCIM News.
\end{IEEEbiography}

\vspace{-0.5cm}

\begin{IEEEbiography}[{\includegraphics[width=1in,height=1.25in,clip,keepaspectratio]{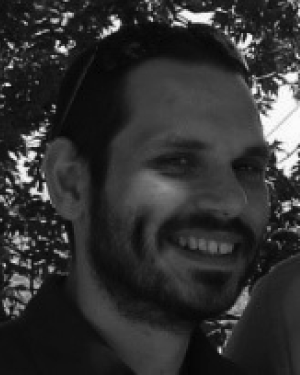}}]{Georgios Mylonas} received the M.Sc., Ph.D., and Diploma degrees from the Department of
Computer Engineering and Informatics, University of Patras. He is currently Principal
Researcher with the Industrial Systems Institute, “Athena” Research and Innovation Center, Patras, Greece. Before joining Athena RC, he was a Senior Researcher at the Computer
Technology Institute and Press “Diophantus.” He has been involved in the AEOLUS, WISEBED,
SmartSantander, OrganiCity, and E2Data EU research projects, focusing on algorithmic and
software issues of sensor networks, the IoT, and smart cities. He has coauthored over 70
journal articles and conference papers. He has coordinated and was the Principal Investigator of the Green Awareness in Action (GAIA) H2020 Project, focusing on sustainability awareness and energy efficiency in the educational sector. His research interests include the areas of the IoT, smart cities, wireless sensor networks, and distributed and pervasive systems.
\end{IEEEbiography}

\vspace{-0.5cm}

\begin{IEEEbiography}[{\includegraphics[width=1in,height=1.25in,clip,keepaspectratio]{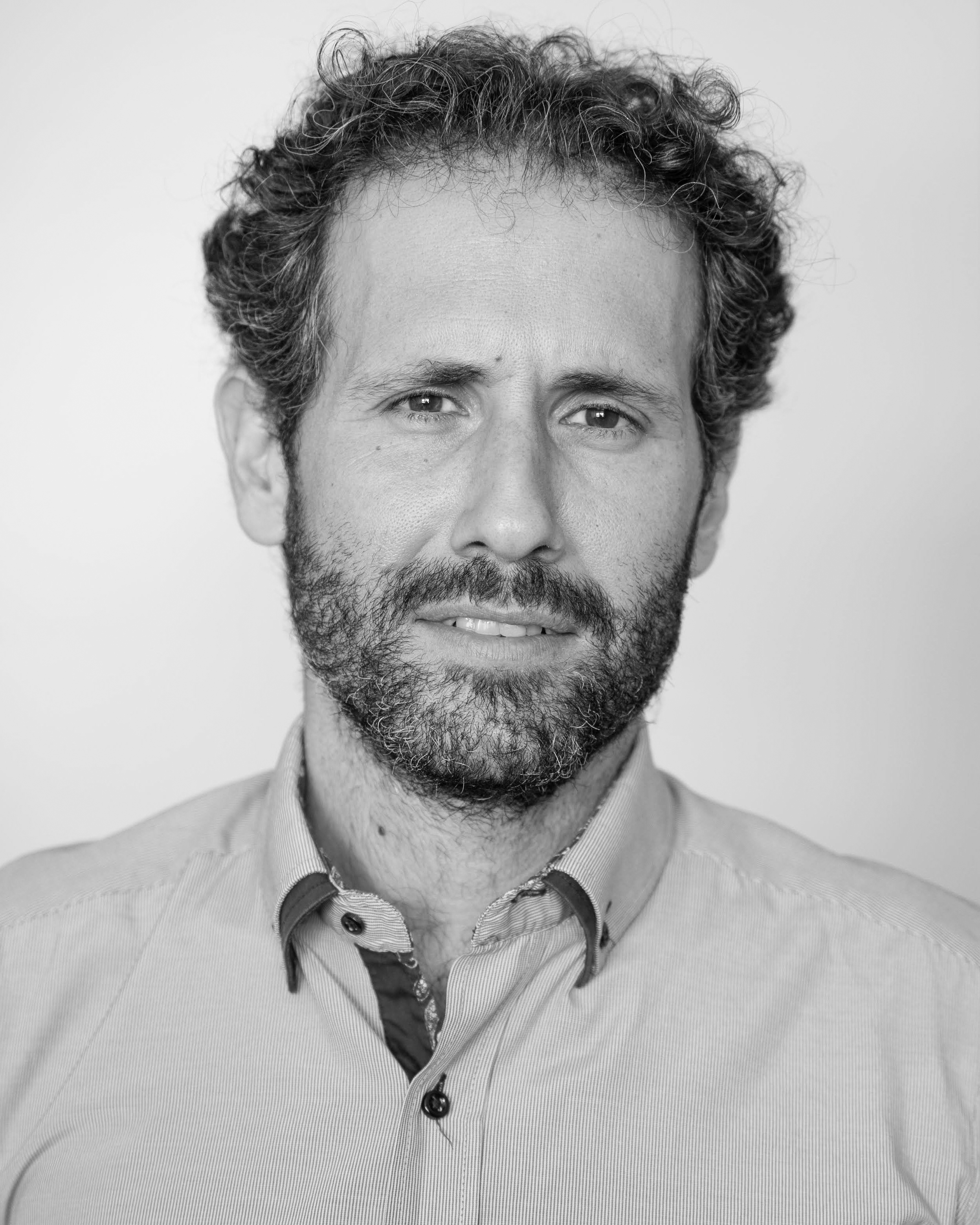}}]{Konstantinos Moustakas}
received the Diploma degree and the PhD in electrical and computer engineering from the Aristotle University of Thessaloniki, Greece, in 2003 and 2007 respectively. During 2007-2011 he served as a post-doctoral research fellow in the Information Technologies Institute, Centre for Research and Technology Hellas. He is currently a Professor at the Electrical and Computer Engineering Department of the University of Patras, Head of the Visualization and Virtual Reality Group, Director of the Wire Communications and Information Technology Laboratory and Director of the MSc Program on Biomedical Engineering of the University of Patras. He serves as an Academic Research Fellow for ISI/Athena research center. His main research interests include virtual, augmented and mixed reality, 3D geometry processing, haptics, virtual physiological human modeling, in- formation visualization, physics-based simulations, computational geometry, computer vision, and stereoscopic image processing. He is a senior member of the IEEE, the IEEE Computer Society and member of Eurographics
\end{IEEEbiography}

\vfill

\end{document}